\documentclass[runningheads,a4paper]{llncs}
\usepackage{graphicx}
\usepackage{pgfplots}
\pgfplotsset{compat=newest}
\usetikzlibrary{plotmarks}
\usepackage[font=small,skip=4pt]{caption}
\usepackage{gensymb}
\usepackage{mathptmx}       
\usepackage{helvet}         
\usepackage{courier}        
\usepackage{type1cm}        
\usepackage{makeidx}         
\usepackage{graphicx}        
\usepackage{multicol}        
\usepackage[bottom]{footmisc}
\usepackage{subfig}
\usepackage{eufrak}
\usepackage{amsmath}
\usepackage{enumitem}
\usepackage{amsfonts}
\usepackage{amssymb}
\usepackage{gensymb}
\usepackage{acronym}
\usepackage{bm}
\usepackage[ruled]{algorithm2e}
\usepackage{fancybox,color}
\DeclareMathAlphabet{\mathcal}{OMS}{cmsy}{m}{n} 

\usepackage{url}
\urldef{\mailsa}\path|{fernando.nobre,christoffer.heckman}@colorado.edu|
\newcommand{\keywords}[1]{\par\addvspace\baselineskip
\noindent\keywordname\enspace\ignorespaces#1}

\DeclareMathOperator*{\argmax}{arg\,max}
\DeclareMathOperator*{\argmin}{arg\,min}

\newacro{SLAM}{Simultaneous Localization and Mapping}
\newacro{MAP}{Maximum-a-Posteriori}
\newacro{IMU}{Inertial Measurement Unit}

\begin{document}

\newlength\figureheight
\newlength\figurewidth

\mainmatter  

\title{FastCal: Robust Online Self-Calibration for Robotic Systems}

\titlerunning{FastCal: Robust Online Self-Calibration}

%
%
\author{Fernando Nobre \and Christoffer R.\ Heckman}%

\authorrunning{Nobre and Heckman}


\institute{Department of Computer Science, University of Colorado, Boulder, CO 80309, USA\\
\mailsa}

%
%

\toctitle{FastCal: Robust Online Self-Calibration for Robotic Systems}
\tocauthor{Nobre and Heckman}
\maketitle

\begin{abstract}
We propose a solution for sensor extrinsic self-calibration with very low time complexity, competitive accuracy and graceful handling of often-avoided corner cases: drift in calibration parameters and unobservable directions in the parameter space. It consists of three main parts: 1) information-theoretic based segment selection for constant-time estimation; 2) observability-aware parameter update through a rank-revealing decomposition of the Fischer information matrix; 3) drift-correcting self-calibration through the time-decay of segments. At the core of our FastCal algorithm is the loosely-coupled formulation for sensor extrinsics calibration and efficient selection of measurements. FastCal runs up to an order of magnitude faster than similar self-calibration algorithms\footnote{camera-to-camera extrinsics, excluding feature-matching and image pre-processing on all comparisons.}, making FastCal ideal for integration into existing, resource-constrained, robotics systems.

\end{abstract}

\section{Introduction} \label{sec:intro}

Autonomous platforms destined for long-term applications equipped with multiple sensors such as cameras and \ac{IMU} have become increasingly ubiquitous. Generally these platforms must undergo sophisticated calibration routines to estimate extrinsic (sensor-to-sensor rigid body transform) parameters to high degrees of certainty before sensor data may be interpreted and fused. Once fielded, calibration parameters are generally fixed for the lifetime of the platform. However, in many applications these platforms experience gradual changes in calibration parameters due to e.g.\ non-rigid mounting, accidental bumps and temperature dilation that can change sensor extrinsic parameters. Self-calibration addresses this by inferring extrinsic parameters pertaining to proprioceptive and exteroceptive sensors without using a known calibration target or a specific calibration routine. The motivation behind self-calibration is to remove the explicit, tedious, and sometimes nearly impossible calibration procedure from robotic applications and to enable robust long-term autonomous operation.
Self-Calibration is an essential part of any long-term robotic system, as such is under constant pressure to increase its accuracy, speed and robustness. A higher speed allows its inclusion into larger systems with extensive subsequent processing (\textit{e.g.} localization, mapping, object/activity recognition, planning) and its deployment in computationally constrained scenarios (\textit{e.g} planetary exploration, embedded systems). A robust self-calibration system should cope with unobservable directions in the parameter space (\textit{e.g.} due to a nonholonomic platform, measurement noise which makes unobservable parameters \textit{appear} observable) and changes and drift in calibration parameters.
\begin{figure}
\centering
\includegraphics[width=\textwidth]
{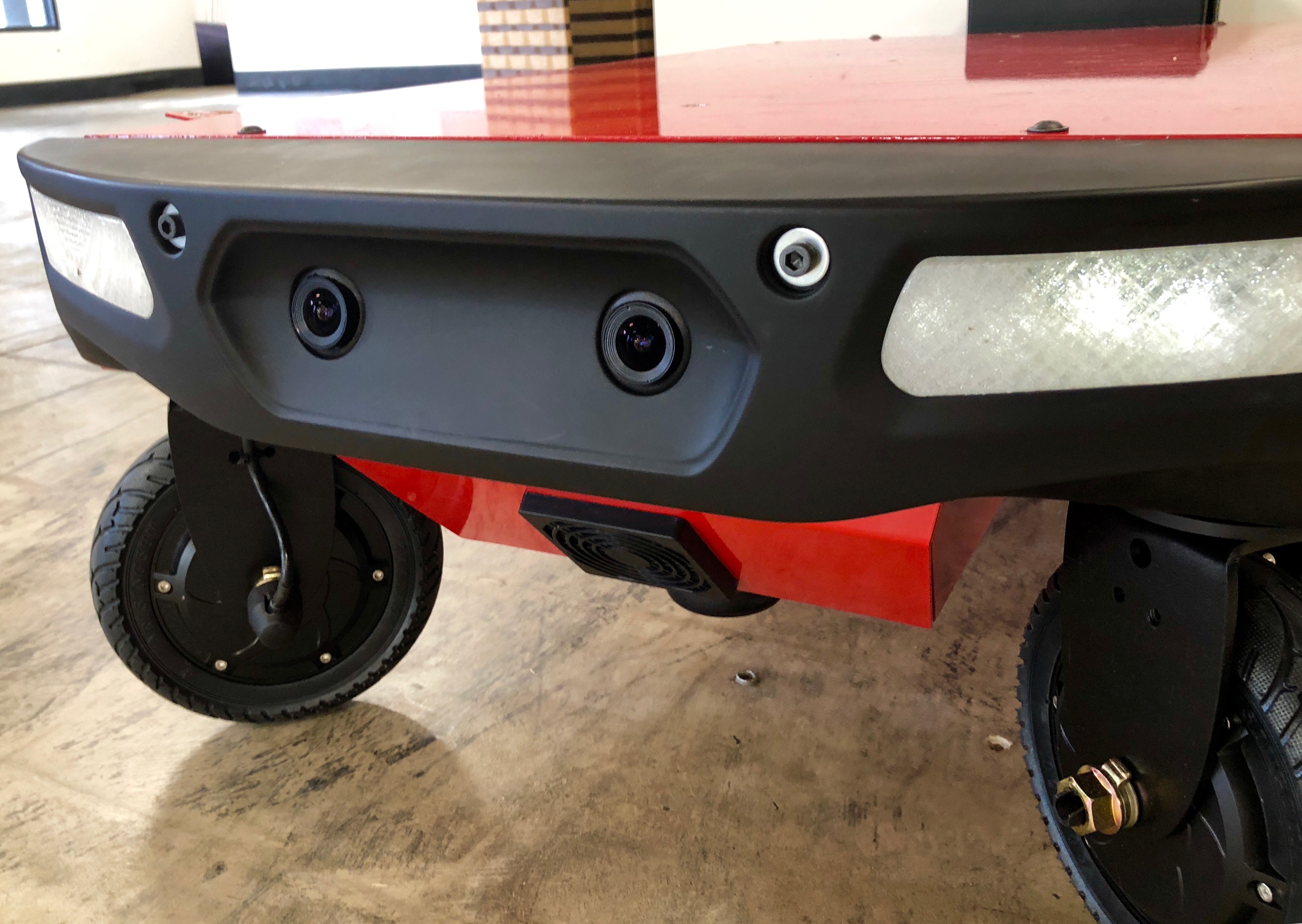}
\caption{Robotic platform equipped with two camera stereo pairs, one in front and one in the back, used for experiments.}
\label{fig:cta_cart}
\end{figure}
In our previous work \cite{NobreISER16} we have partially addressed constant-time self-calibration using a priority queue and informative segments. \cite{nobre2017icra} addresses drift and slow changes in calibration parameters by attempting to regress the change point and retroactively correcting the state estimates and in \cite{NobreISRR17} we propose an observability aware framework capable of updating only the observable directions of the parameter space, even in the presence of noisy measurements, which is then used in a reinforcement-learning framework to learn informative motions to be suggested to a human operator. In this paper we leverage some individual aspects of previous works and propose significant novel contributions:
\begin{enumerate}[label=\textbf{\arabic*}]
\item{A novel formulation for regressing extrinsic calibration parameters which is suitable for integration with any existing \ac{SLAM} system, while being considerably faster than the system used in \cite{NobreISER16} due to a loosely coupled formulation which optimizes over relative poses instead of jointly over raw sensor measurements. We also propose a novel criteria for adding informative segments to the estimation queue which minimizes the number of times the entire segment queue needs to be optimized.}
\item{We handle intrinsically degenerate scenarios in noisy nonholonomic systems by only updating the observable directions of the parameter space, demonstrating the necessity and usefulness in a real-world robotic application.}
\item{Slow changes over arbitrary periods of time are handled by continuously renewing the segment queue by use of time-decay on measurements, this approach is shown to be much more efficient and robust compared to \cite{nobre2017icra}, at the cost of immediate and local accuracy. }
\item{Integration of the proposed self-calibration system into a real-world robotic platform operating in challenging environments for extended periods of time\footnote{3 weeks of operation, approximately 170km driven.}. }
\end{enumerate}

\subsection{Related Work}\label{subsec:related}
Most current techniques for vision-aided inertial navigation use filtering approaches \cite{Jones:2011fr,Kelly:2011bw,Mourikis:2007dm} or a smoothing formulation. In either case the estimation is made constant-time by rolling past information into a prior distribution. Filtering methods present the significant drawback of introducing inconsistencies due to linearization errors of past measurements which cannot be corrected post hoc, particularly troublesome for non-linear camera models. Some recent work has tackled these inconsistencies; see, e.g.\ \cite{Li:2013co,Hesch:2013jf,Civera:2009kc,Li:2014jx}. The state-of-the-art includes methods to estimate poses and landmarks along with calibration parameters, but these approaches do not output the marginals for the calibration parameters, which are desirable for long-term autonomy applications.

The use of a known calibration pattern such as a checkerboard coupled with nonlinear regression has become the most popular method for camera calibration in computer vision during the last decade; it has been deployed both for intrinsic camera calibration \cite{sturm1999plane} and extrinsic calibration between heterogeneous sensors \cite{zhang2004extrinsic}. While being relatively efficient, this procedure still requires expert knowledge to reach a discerning level of accuracy. It can also be quite inconvenient on a mobile platform requiring frequent recalibration (e.g experimental platforms which undergo constant sensor changes). In an effort to automate the process in the context of mobile robotics, several authors have included the calibration problem in a state-space estimation framework, either with filtering \cite{martinelli2006automatic} or smoothing \cite{kummerle2011simultaneous} techniques. Filtering techniques based on the Kalman filter are appealing due to their inherently online nature. However, in case of nonlinear systems, smoothing techniques based on iterative optimization can be superior in terms of accuracy \cite{strasdat2010real}.

Our approach does not rely on formal observability analyses to identify degenerate paths of the calibration run as in \cite{brookshire2013extrinsic}, since these approaches still expect non-degenerate excitations.

A last class of methods relies on an energy function to be minimized. For instance, Levinson and Thrun \cite{levinson2014unsupervised} have defined an energy function based on surfaces and Sheehan et al.\ \cite{sheehan2012self} on an information theoretic quantity measuring point cloud quality.

\section{Methodology} \label{sec:methodology}

It is common for robotic platforms to have multiple sensors, such as cameras, wheel encoders, \ac{IMU}. This creates the need to obtain the relative rigid body transform between sensors so that a fused position estimate may be obtained. This is what we refer to as \textit{calibration parameters} in this work, represented by $\Theta$. There are other calibration parameters which can be estimated, such as camera intrinsics (\textit{e.g.}focal length, center point, distortion parameters) but these have been found to not vary considerably even in long term operation. Sensor extrinsics however, change frequently (Section \ref{sec:results}) and have considerable impact on the resulting position estimation. For these reasons we focus on estimating sensor extrinsics. 
\begin{figure}
\subfloat[Tightly coupled problem.\label{subfig-1:dummy1}]{
\includegraphics[width=0.5\textwidth,height=110pt]{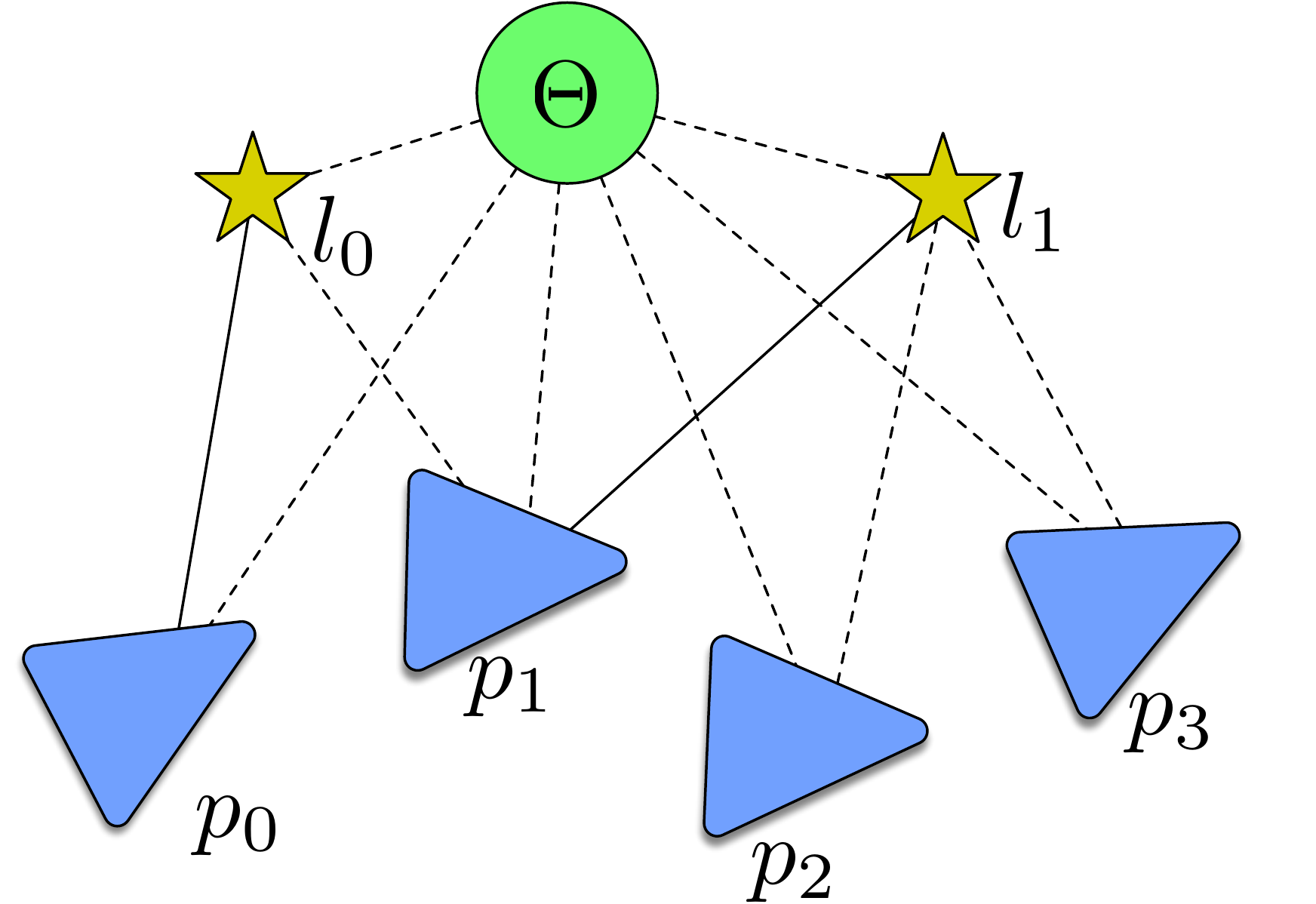}
}
\subfloat[Loosely coupled problem.\label{subfig-1:dummy1}]{
\includegraphics[width=0.5\textwidth,height=90pt]{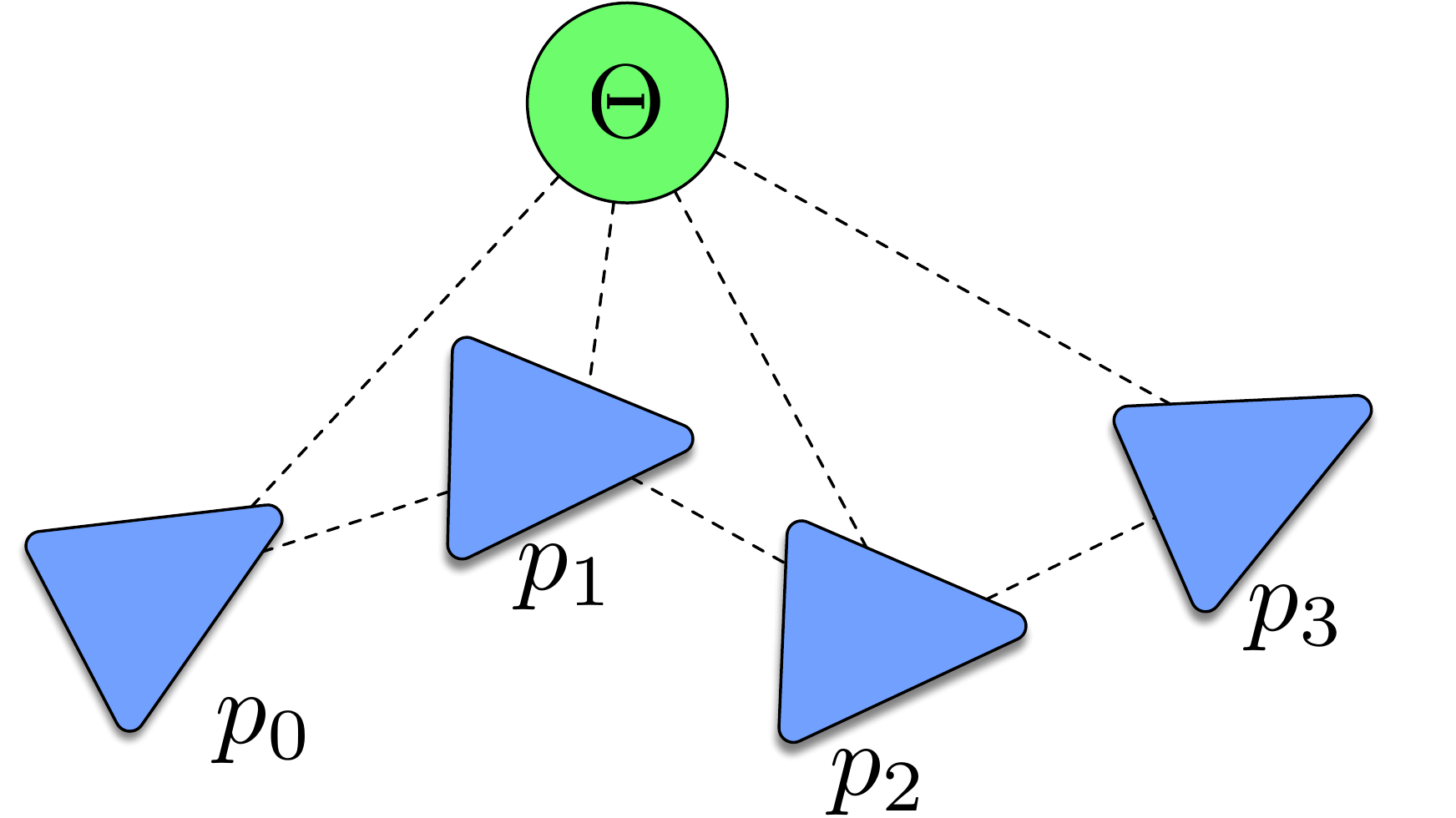}
}
\hfill
\subfloat[System diagram.\label{subfig-1:dummy2}]{
\includegraphics[width=1.0\textwidth]{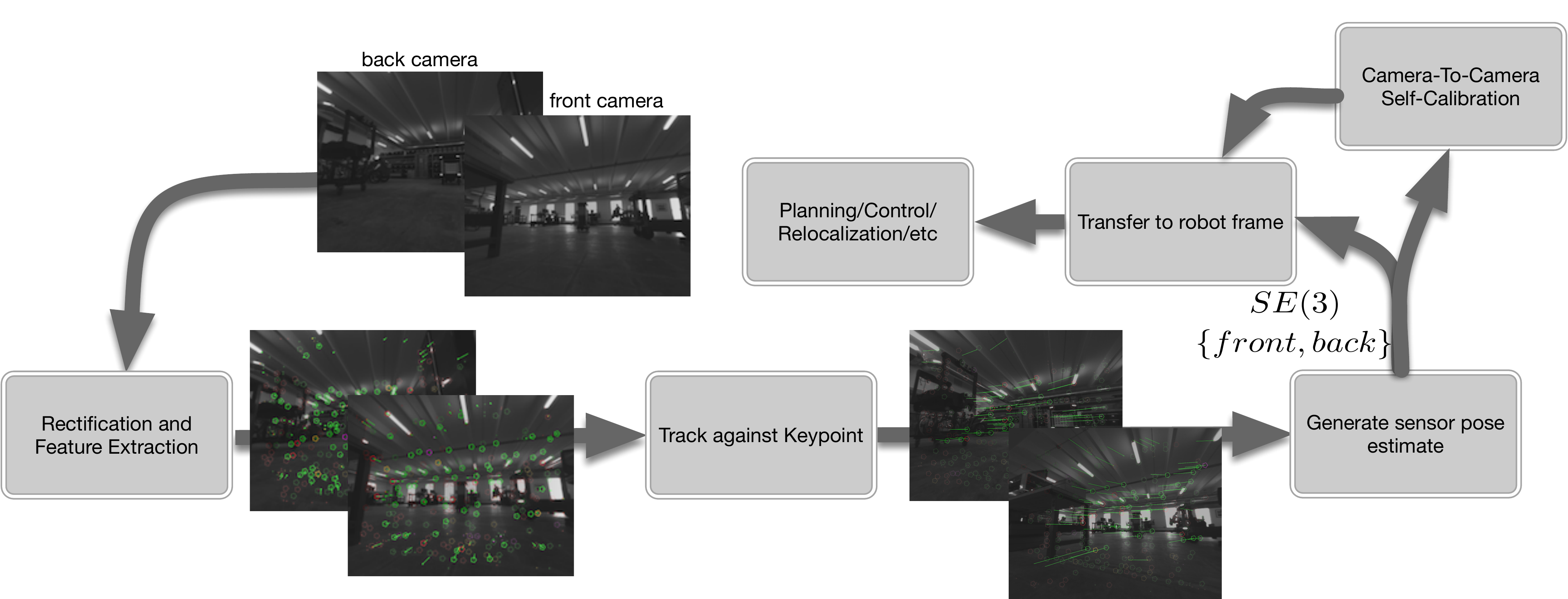}
}
\caption{(a) Landmark measurements are used to jointly estimate camera poses and camera-to-camera extrinsics ($\Theta$). (b) Camera poses are estimated independently; The calibration parameter $\Theta$ is estimated in a second step. (c) System diagram of a typical SLAM system, the input is each camera image, and the self-calibration block integrates easily into any existing odometry pipeline by subscribing to the existing odometry block.}
\label{fig:blocks}
\end{figure}
For camera intrinsic self-calibration refer to \cite{NobreISER16,Keivan:2014bl}. The proposed FastCal algorithm can be divided into three major components, summarized in algorithm \ref{algo:fast_cal}: 1) Selecting informative segments so as to bound the computation time. 2) Updating only the observable directions of the parameter space. 3) Considering drift in the calibration parameters over time by time-decaying measurements. Figure \ref{fig:blocks} shows the calibration problem posed as a factor graph, both as a tightly couple problem where the poses, landmarks and calibration parameters are estimated jointly and as a loosely coupled problem where the individual sensor pose-graphs are estimated independently, and the calibration parameters are obtained in a subsequent optimization. The block diagram for a typical SLAM system is also shown, where there is usually an odometry node which provides relative poses. We also focus on creating an algorithm which has as few tuning parameters as possible; the parameters for FastCal are summarized in table \ref{tab:parameters}, the parameters in bold are the ones with highest impact on the performance of the algorithm.

\vspace{8pt}
\begin{algorithm}[H]
 \KwData{relative pose measurements for each of the $N$ sensors; reference sensor $N_{ref}$, initial guess on sensor placement.}
 \KwResult{sensor-to-sensor extrinsic $SE(3)$: $\Theta$}
 Initialize $\Theta \leftarrow$ initial guess\;
 \eIf{$num\_measurements(\mathcal{D}^{candidate}) < \theta_{meas}$}{
  $\mathcal{D}^{candidate} \leftarrow $ new measurements\;
  }{
    Estimate $\Theta | \mathcal{D}^{candidate} $ according to \eqref{eq:theta_argmax}\;
    \eIf{$num\_segments(\mathcal{D}^{info}) < \theta_{pq}$}{
		$\mathcal{D}^{info} \leftarrow \mathcal{D}^{candidate}$\;
  	}{
     Check if $\mathcal{D}^{candidate}$ should be swapped into $\mathcal{D}^{info}$, according to \ref{subsec:priority_queue} \;
     \If{$\mathcal{D}^{info} \leftarrow \mathcal{D}^{candidate}$}{
     	Estimate $\Theta | \mathcal{D}^{info} $ according to \eqref{eq:svd_update} with TSVD\;
     }
    }
  }
  \For{$i\leftarrow 0$ \KwTo $num\_segments(\mathcal{D}^{info})$}{
  	\If{Time Decay for $\mathcal{D}^{candidate}_{i}$ according to \eqref{eq:time_decay} $< 0.001$}{
    	remove $\mathcal{D}^{candidate}_{i}$ from $\mathcal{D}^{info}$
    }
  }
 
 \caption{FastCal Algorithm}
 \label{algo:fast_cal}
\end{algorithm}
\vspace{8pt}

\begin{table}
\caption {FastCal Parameters} \label{tab:parameters} 
\begin{center}
\begin{tabular}{ c | c | c }
  Function & Symbol & Default Value \\ \hline
  \textbf{TSVD threshold} & $\theta_{\epsilon_{svd}}$ & $0.1$ \\
  \textbf{Maximum entropy} & $\theta_{\Sigma_{max}}$ & $15$ \\
  \textbf{Number of segments in priority queue} & $\theta_{pq}$ & $10$ \\
  \textbf{Number of measurements in each candidate} & $\theta_{meas}$ & $10$ \\
  \textit{Keyframing translation [m]} & $\theta_{kf_{trans}}$ & $0.15$ \\
  \textit{Keyframing rotation [rad]} & $\theta_{kf_{rot}}$ & $0.1745$  \\
  \textit{Time decay} & $\theta_{\lambda}$ & $0.04$ \\
  \textit{Number of consecutive estimates at the same value} & $\theta_{same}$ & $3$ \\
  \textit{Min total update} & $\theta_{min\_update}$ & $0.008$ \\
\end{tabular}
\end{center}
\end{table}

\subsection{Problem Formulation}
We focus on estimating the sensor-to-sensor SE(3) rigid body transform between two cameras with no co-visible features.
The calibration problem can be framed as an optimization problem in a Bayesian estimation framework, by including the calibration parameters in the standard \ac{SLAM} formulation:
\begin{equation}
\hat{\mu}_{\Theta\mathcal{X}\mathcal{L}} = \argmax_{\Theta\mathcal{X}\mathcal{L}}p(\Theta, \mathcal{X}, \mathcal{L}|\mathcal{Z}) = \argmin_{\Theta\mathcal{X}\mathcal{L}}-\text{log}\,p(\Theta, \mathcal{X}, \mathcal{L}|\mathcal{Z});
\label{eq:normal_means}
\end{equation}
Where the estimated parameter $\mu$ contains the robot pose ($\mathcal{X}$), landmarks ($\mathcal{L}$) and calibration ($\Theta$) parameters. $\mathcal{Z}$ in this context are the sensor measurements, such as landmark observations. The advantages of solving the problem in this formulation is leveraging joint information from all measurements, at the price of higher computational complexity since we must solve a larger system comprised of $N$ poses, $M$ landmarks in addition to the calibration parameters.

Alternatively, we can leverage the fact that most robotics systems already estimate the reduced state:

\begin{equation}
\hat{\mu}_{\mathcal{X}\mathcal{L}} = \argmax_{\mathcal{X}\mathcal{L}}p(\mathcal{X}, \mathcal{L}|\mathcal{Z}) = \argmin_{\mathcal{X}\mathcal{L}}-\text{log}\,p(\mathcal{X}, \mathcal{L}|\mathcal{Z});
\label{eq:normal_no_theta}
\end{equation}

for each camera, where $\mathcal{X} = [\mathbf{x}_{s1}, \mathbf{x}_{s2}, ...,\mathbf{x}_{sn}]$ the world position for $N$ sensors at time $t$. Finding the sensor-to-sensor extrinsics can then be posed as an alignment problem, using $\mathcal{X}$ as the measurement instead of the landmark observations as in \eqref{eq:normal_means}.

\begin{equation}
\hat{\Theta} = \argmax_{\Theta}p(\Theta | \mathcal{X}) = \argmin_{\Theta}-\text{log}\,p(\Theta | \mathcal{X});
\label{eq:theta_argmax}
\end{equation}

This formulation has a few advantages: the least squares solution to \eqref{eq:theta_argmax} is simply a $6\times6$ system (for a single pair of sensors) which can be solved very efficiently and allows for the use of more informative decompositions as will be discussed in \ref{subsec:observability}. This formulation also allows for easy integration into an existing \ac{SLAM} system which already provides the independent sensor position estimates; All that needs to be done is subscribe to the camera positions being estimated and efficiently solve \eqref{eq:theta_argmax}.

\subsection{Informative Segment Selection}\label{subsec:priority_queue}
 We want to benefit from the robustness
of offline methods, along with the possibility to deploy
our algorithm in an online and long-term setting. To reach
this goal, we process small batches of data sequentially and
decide to keep them based on their utility for the calibration.
Each new batch is merged to the old ones to refine
our knowledge about the calibration parameters until we
reach a satisfactory level of confidence. This notion was introduced in \cite{Keivan:2014bl} and extended to multiple sensors in \cite{NobreISER16}. We briefly explain the informative segment selection into a priority queue, to then introduce our novel metric on adding new batches of measurements. We define the current set of data samples in a priority queue as $\mathcal{D}^{info} = \{\mathbf{x}_1,...,\mathbf{x}_n\}$ which has led to the posterior marginal density in equation \eqref{eq:theta_argmax}, associated to the random variable $\hat{\Theta} | \mathcal{D}^{info}$. The set, $\mathcal{D}^{info}$, contains the current
informative measurements for the calibration variable $\Theta$. Our
sensors, $S$, continuously stream new data which are used for estimating their relative positions, that we then accumulate in another batch of size $\Delta N$ denoted by $\mathcal{D}^{candidate} = \{\mathbf{x}_{n+1},...,\mathbf{x}_{n + \Delta N}\}$.  Intuitively, if the measurements in $\mathcal{D}^{candidate}$ are similar to those in $\mathcal{D}^{info}$, we are not
really improving our knowledge about $\Theta$ and we can
safely discard $\mathcal{D}^{candidate}$ to keep the computation tractable. There are multiple ways of evaluating the usefulness of $\mathcal{D}^{candidate}$: Given the covariance of the estimated calibration parameters $\Sigma_{\Theta}$, which can be obtained quickly from the solution to \eqref{eq:theta_argmax} by inverting the $6\times6$ information matrix, the entropy of the distribution is then given by $h = \frac{1}{2}\text{ln}|2\pi e \Sigma_{\Theta}|$, where the bars denote the matrix determinant. In \cite{NobreISER16}, the update criteria was if the entropy $h_{candidate}$ associated to the measurements $\mathcal{D}^{candidate}$ was smaller than the worst scoring batch in $\mathcal{D}^{info}$ by a certain margin, the segment was swapped into the informative segment queue and a new estimate for $\Theta$ was obtained by optimizing over all the measurements in the queue. This approach performs well, as shown in \cite{Keivan:2014bl,NobreISER16}, however it causes an excessive number of estimations of the entire priority queue, every time a new candidate segment is swapped in.

\begin{figure}[h]  
\centering 
\resizebox {\textwidth}{!} {
%
%
\definecolor{mycolor1}{rgb}{1.00000,1.00000,0.00000}%
\definecolor{mycolor2}{rgb}{0.20810,0.16630,0.52920}%
\begin{tikzpicture}

\begin{axis}[%
width=2.157in,
height=1.645in,
at={(0.838in,0.424in)},
scale only axis,
bar shift auto,
xmin=0,
xmax=8,
xtick={1, 2, 3, 4, 5, 6},
xlabel style={font=\color{white!15!black}},
xlabel={time},
ymin=0,
ymax=35,
ylabel style={font=\color{white!15!black}},
ylabel={entropy},
axis background/.style={fill=white},
legend style={legend cell align=left, align=left, draw=white!15!black}
]
\addplot[ybar, bar width=0.8, fill=mycolor1, draw=black, area legend] table[row sep=crcr] {%
1	34\\
2	29\\
3	22\\
4	16\\
5	25\\
6	28\\
};
\addplot[forget plot, color=white!15!black] table[row sep=crcr] {%
0	0\\
8	0\\
};
\addlegendentry{candidates}

\end{axis}

\begin{axis}[%
width=2.157in,
height=1.645in,
at={(3.676in,0.424in)},
scale only axis,
bar shift auto,
xmin=0,
xmax=4,
xtick={1, 2, 3},
xlabel style={font=\color{white!15!black}},
xlabel={segment},
ymin=0,
ymax=35,
ylabel style={font=\color{white!15!black}},
ylabel={entropy},
axis background/.style={fill=white},
legend style={legend cell align=left, align=left, draw=white!15!black}
]
\addplot[ybar, bar width=0.8, fill=mycolor2, draw=black, area legend] table[row sep=crcr] {%
1	10\\
2	30\\
3	17\\
};
\addplot[forget plot, color=white!15!black] table[row sep=crcr] {%
0	0\\
4	0\\
};
\addlegendentry{priority queue}

\end{axis}
\end{tikzpicture}%
}
\caption{Left: Entropies of rolling candidate window over time. Candidate window at time $t=2$ could be swapped into the priority queue in place of segment 2, however by waiting until time $t=5$ we can instead swap in the candidate window at time $t=4$, with much more information content. Right: Entropy of three informative batches in the priority queue.}
\label{fig:pq_entropies}
\end{figure}
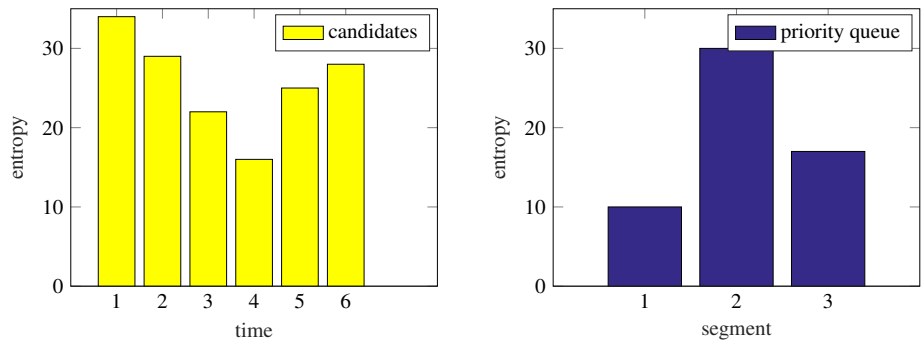
We propose a different metric on adding batches to the priority queue: a candidate measurement batch is swapped into the queue iff. its entropy beats the worst scoring segment in the priority queue \textit{and} the rolling candidate window has achieved a local entropy minimum. This ensures that we dont needlessly swap in segments to the priority queue and ensures that we are adding the best possible local window to the queue, not just the one that beat the worst scoring segment. Figure \ref{fig:pq_entropies} shows on the left the entropy for a rolling window of $N=10$ candidate measurements over time, on the right the entropies for the three segments in the priority queue. In the previous formulation the candidate window at time $t=2$ has an entropy of 27 which beats segment 2 in the priority queue (with an entropy of 30) and would be swapped in. However by holding off on adding the candidate window to the priority queue, at time $t=5$ we notice that the candidate entropy is increasing and as such we have reached a local minimum at $t=4$, which is then swapped into the priority queue. This approach reduces the priority queue entropy much faster and leads to quicker convergence on $\Theta$, as well as greatly reducing the number times the priority queue needs to be estimated.

The measurements which are added to each segment are the relative pose measurements generated independently for each sensor (each camera in our case) as shown in Figure \ref{fig:blocks}. Most information based-approaches add every measurement to the candidate window, and re-estimate the calibration parameter $\Theta | \mathcal{D}^{candidate}$, however since most systems can generate relative pose measurements at over 20Hz, the information gain from each new individual measurement is small and results in several evaluations of $\mathcal{D}^{candidate}$. We aim to reduce the number of candidate window evaluations. Instead of adding each new measurement we take a keyframing based approach to self-calibration and accumulate relative pose measurements until the total relative transform reaches a threshold. In our case we use $\theta_{kf_{trans}} = 0.15 m$ and $\theta_{kf_{rot}} = 0.1745 rad$. This results in both much faster convergence and reduces the number of times the candidate window and priority queue need to be estimated.

\subsection{Observability-Aware Estimation}\label{subsec:observability}
We wish to reliably estimate the sensor extrinsics even in nonholonomic platforms which can never excite all degrees of freedom, and thus will not render the full calibration parameter space observable. We leverage our insights from \cite{NobreISRR17} to use the Truncated SVD (TSVD) decomposition of the Fischer Information Matrix (FIM) and determine the observable directions in the parameter space, and only update those, even in the presence of noise. This is essential for platforms which, for example, only move on a planar surface and never excite roll, pitch or translation along the direction normal to the ground plane. In these cases the system \eqref{eq:theta_argmax} is ill conditioned and we have to either manually regularize the unobservable directions in order to make the system well conditioned or automatically detect the observable components and only update those directions. This is the approach we take in this work; We argue that manually regularizing explicitly throws away information which could be useful and implicitly biases the solution. Furthermore, by adopting the simplified parameter space, we can perform the SVD decomposition on the FIM very quickly, since it is of reduced size ($6\times 6$  for a pair of sensors). The least squares solution to \eqref{eq:theta_argmax} is

\begin{equation}
(\mathbf{J}^{T}\mathbf{G}^{-1}\mathbf{J})\delta\hat{\Theta} = -\mathbf{J}^{T}\mathbf{G}^{-1}\mathbf{r}(\hat{\Theta}),
\label{eq:normal}
\end{equation}
Where $J$ is the Jacobian matrix, $G$ the measurement covariance matrix obtained from the first step and $r$ the residual.
There exists a solution to Eq.\ \eqref{eq:normal} iff the FIM is invertible, i.e.\ it
is of full rank. The link between the rank of the FIM and observability of the parameters being estimated is well established in \cite{jauffret2007observability}. A singular FIM corresponds to some unobservable directions in the parameter space given the current set of observations. Classical observability analysis, for example the method of Hermann and Krener \cite{hermann1977nonlinear}, proves structural observability---that there exists some dataset for which the parameters are observable---but it does not guarantee that the parameters are observable for any dataset.

Using singular-value decomposition (SVD) on the FIM we can identify a numerically rank-deficient matrix by analyzing its singular values and consequently the numerical observability of the system \cite{hansen1998rank}. The \textit{numerical rank} $r$ of a matrix is defined as the index of the smallest singular value $\sigma_r$ which is larger than a pre-defined tolerance $\theta_{\epsilon_{svd}}$,
$r = \argmax_{i}\sigma_i \geq \theta_{\epsilon_{svd}}$. When the noise affecting the matrix entries has the same scale (by using column or row scaling) then the numerical rank can be determined by the singular values. The scaling matrix $S$ can be computed as:
\begin{equation}
S = \text{diag}\left\{\frac{1}{||J(:, 1)||},...,\frac{1}{||J(:, n)||}\right\}
\end{equation}
Where $||J(:, n)||$ denotes the column norm of the Jacobian matrix, for column $n$.
Specifically we decompose the error covariance matrix $\mathbf{G}$ from Eq.\ \eqref{eq:normal} into its square root form by using Cholesky decomposition, $\mathbf{G}^{-1} = \mathbf{L}^{T}\mathbf{L}$, we can re-write Eq\ \eqref{eq:normal} in standard form:$
(\mathbf{L}\mathbf{J})^{T}(\mathbf{L}\mathbf{J}) \delta\hat{\Theta} = -(\mathbf{L}\mathbf{J})^{T}\mathbf{L} \mathbf{r}(\hat{\Theta})$ which are the normal equations for the linear system $(\mathbf{L}\mathbf{H})\delta\hat{\Theta} = -\mathbf{L} \mathbf{r}(\hat{\Theta})$. Thus we can directly use a \textit{rank-revealing} decomposition to estimate the numerical rank of the FIM, and consequently the numerical observability of the system.  Let $(\mathbf{L}\mathbf{J})$ be a $m \times n$ matrix with the following SVD decomposition:
\begin{equation}
\mathbf{LJ} = \mathbf{US}\mathbf{V^T},
\label{eq:SVD}
\end{equation}
where $\mathbf{U}$ is $m \times n$ and orthogonal, $\mathbf{S} = diag(\varsigma_1,...,\varsigma_i)$ the singular values and $\mathbf{V}$ an $n \times n$ matrix, also orthogonal. From Eq.\ \eqref{eq:SVD} and the orthogonality of $\mathbf{U}$ and $\mathbf{V}$ we can solve (\ref{eq:normal}) as
\begin{equation}
 \delta\hat{\Theta} = -\mathbf{V}\mathbf{S^{-1}}\mathbf{U}\mathbf{L}\mathbf{r}(\hat{\Theta}),
 \label{eq:svd}
\end{equation}
Specifically, according to \cite{hansen1998rank}, we can efficiently obtain the update as: 
\begin{equation}
\delta\hat{\Theta} = \sum_{i=1}^{r_\epsilon}\frac{\mathbf{u}_{i}^{T}\mathbf{r}_{\Theta}}{\varsigma_i}\mathbf{v}_i,
\label{eq:svd_update}
\end{equation}
\noindent where $\mathbf{u}$ and $\mathbf{v}$ are the colmn vectors of $\mathbf{U}$ and $\mathbf{V}$.
This allows us to only update the observable directions of the parameter space and maintain the other directions at their initial value. Establishing the value of $\epsilon$ to use is specific to the amount of noise expected in the measurements and is treated in Section \ref{sec:results}.

\subsection{Drift Correction}\label{drift}
In order to correct for the inevitable drift over time on the calibration parameters due to physical shocks, maintenance, etc. We adopt a simpler strategy than what was used in \cite{nobre2017icra}; We associate a exponential time-decay with each batch in the priority queue $\mathcal{D}ˆ{info}$. This is done by using a exponential distribution
\begin{equation}
p(t;\lambda) = \lambda e^{-\lambda t}, t \in [0, \infty]
\label{eq:time_decay}
\end{equation}
Which has an expectation $E = \lambda^{-1}$, so we see that the parameter $\lambda$ in \eqref{eq:time_decay} encodes the expected time that a set of measurements remains informative. There are several potential methods for selecting this parameter in practice, among them \textit{class-conditional learning} could use machine learning techniques to learn class-conditional decay rates for certain environments (\textit{e.g.} a warehouse where things move around a lot vs. a building where things rarely change). In this paper we set the decay rate $\lambda$ to a value empirically shown to balance continuously refreshing segments and number of priority queue estimations.

\section{Experiments and Results} \label{sec:results}

\begin{figure}[h!]
\centering
\resizebox {\textwidth}{!} {
	\input{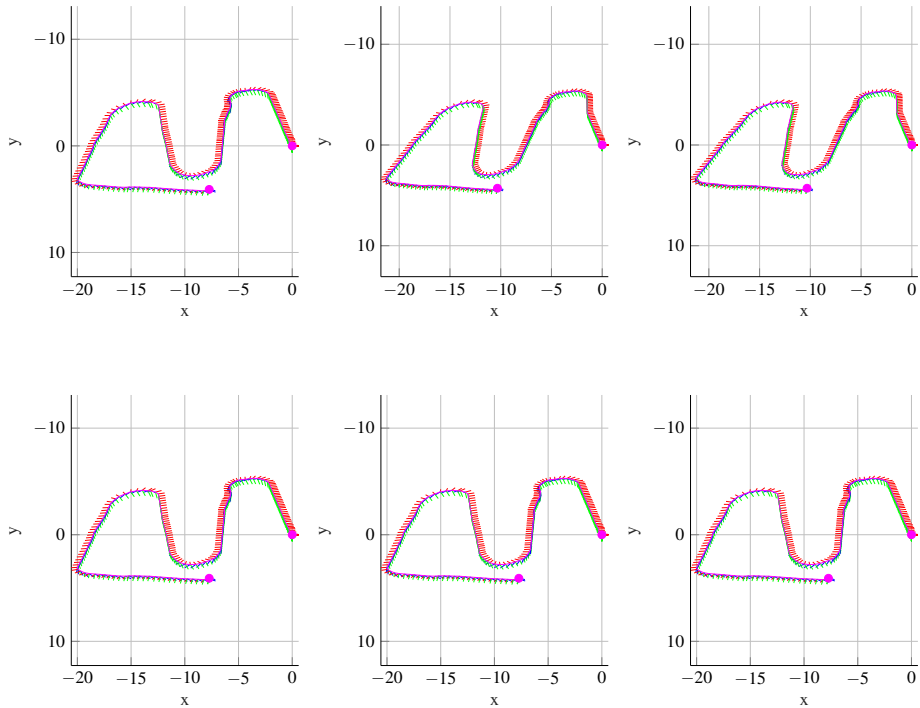}
}
\caption[Trajectory plots.]{Same route performed over the span of multiple weeks. Top row: no self-calibration, using initial offline calibration only. Bottom row: FastCal enabled. Each column represents a day in which that route was performed. Columns are approximately 1 week apart.}
\label{fig:poses_fastcal}
\end{figure}

Our experimental setup consists of a robotic platform (Figure \ref{fig:cta_cart}) designed to transport material autonomously in warehouses, deployed in diverse real-world scenarios. This is an example of a challenging long term autonomy deployment in adverse industrial scenarios, with constrained resources and subject to physical impacts. It proves to be an excellent test-bed for the proposed FastCal algorithm which integrates into the SLAM module seamlessly by subscribing to sensor pose updates and registering a calibration update callback.

\begin{figure}[h!]
\centering
\resizebox {\textwidth}{!} {
	\input{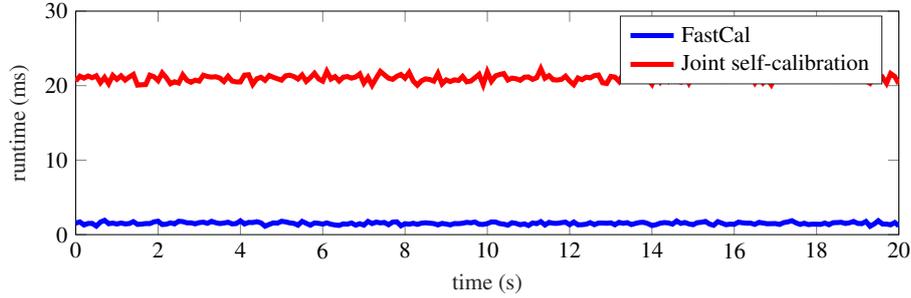}
}
\caption[Timing comparison.]{Timing comparison of the proposed FastCal algorithm vs a reference implementation of tightly coupled, priority queue based self-calibration, for the same size priority queue, candidate segment, number of iterations and fair parameter settings.}
\label{fig:timing}
\end{figure}

The robotic platform is equipped with two pairs of global shutter stereo camera pairs, one facing forward and one facing backward, with no overlapping field of view. We focus on estimating the front-to-back camera extrinsics, given an initial rough guess obtained with a measuring tape. We implement the proposed FastCal algorithm in C++, utilizing the parameters defined in Table \ref{tab:parameters}. We wish to assess how often and by how much the extrinsics parameters actually change in practice. For that end we run FastCal on the robot for a period of 14 days. In this time the robot was subject to cargo loads up to 150kg, was transported in a truck, had it's front bumper removed and re-attached due to maintenance reasons and finally suffered one accidental head-first collision while on joystick mode. These circumstances provide valuable data points on the usefulness and necessity of a robust self-calibration algorithm.

\begin{figure}[h!]
\centering
\resizebox {\textwidth}{!} {
	\input{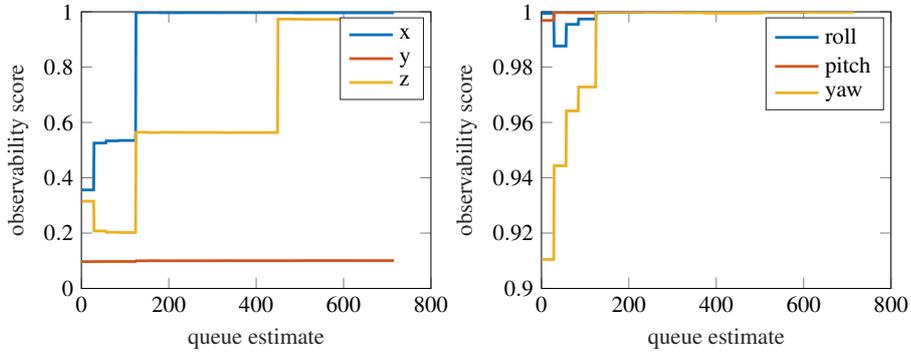}
}
\caption[Observability.]{Observability score for each direction of the calibration parameter (SE(3) transform between cameras). Note that due to the planar motion, the $y$ component is completely unobservable, even in the presence of noise; the unobservable direction remains clamped at its original value as a natural consequence of FastCal, with no need for explicit regularization.}
\label{fig:observability_fastcal}
\end{figure}

\begin{figure}[h!]
\centering
\resizebox {\textwidth}{!} {
	\input{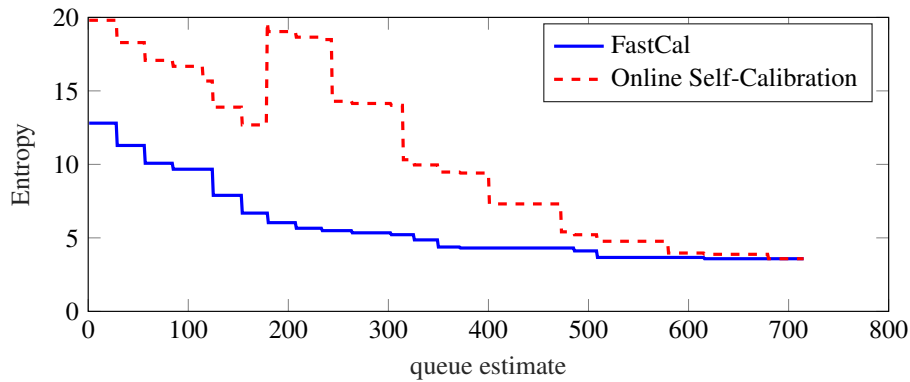}
}
\caption[Entropies comparison.]{Priority Queue entropy comparison of the proposed FastCal algorithm vs a implementation of \cite{NobreISER16}. The novel criteria on adding segments to the queue, coupled with the keyframing on measurements results in a much faster reduction in entropy and thus convergence on the calibration parameters.}
\label{fig:timing}
\end{figure}

\section{Conclusion} \label{sec:conclusion}
In this paper we presented a low time complexity extrinsics self-calibration algorithm that uses an information theoretic measure to add only the most locally informative measurement batches to the estimation queue, the work from \cite{NobreISER16} is improved upon by leveraging novel queue update techniques which drastically reduce the time it takes to converge on the calibration parameters. We have leveraged truncated SVD/QR decomposition to deal with unobservable directions and operate on nonholonomic platforms, similar to the algorithm presented in \cite{NobreISRR17}, however not applied to reinforcement learning and user motion suggestions, but to calibrating nonholonomic robotic platforms. Finally, we have incorporated time-decay as a mechanism to address drift in calibration parameters. We draw from the lessons learned from \cite{nobre2017icra} to implement a simpler, more effective mechanism of dealing with drift. We have evaluated the proposed system in a variety of long term scenarios spanning 15 days and over 120km driven and shown that it can be used as a low-overhead add-on to existing perception systems, while achieving similar accuracy to off-line calibration techniques. Our principal goal was to develop a robust, low time complexity self-calibration algorithm for sensor extrinsics, and show its usefulness in practical long term robotic applications. We have shown that through the decoupling of the estimation problem into two steps, and selectively adding new segments to the priority queue we are able to achieve robust and accurate calibration results with minimal compute overhead. An argument could be made that calibration parameters do not vary enough to justify the addition of self-calibration, but we argue that for true long-term autonomy applications, robust self-calibration is essential, as even in relatively short experiments there was significant change in sensor extrinsics.

\section{Acknowledgments}
The authors gratefully acknowledge support from DARPA award no.\ N65236-16-1-1000.

\bibliographystyle{splncs}
\bibliography{Refs}

\begin{thebibliography}{10}

\bibitem{NobreISER16}
Nobre, F., Heckman, C., Sibley, G.:
\newblock Multi-sensor slam with online self-calibration and change detection.
\newblock In: International Symposium on Experimental Robotics (ISER). (2016)

\bibitem{nobre2017icra}
Nobre, F., Kasper, M., Heckman, C.:
\newblock Drift-correcting self-calibration for visual-inertial slam.
\newblock In: International Conference on Robotics and Automation (ICRA), IEEE
  (2017)

\bibitem{NobreISRR17}
Nobre, F., Heckman, C.:
\newblock Reinforcement learning for assisted visual-inertial robotic
  calibration.
\newblock In: International Symposium on Robotics Research (ISRR). (2017)

\bibitem{Jones:2011fr}
Jones, E.S., Soatto, S.:
\newblock {Visual-inertial navigation, mapping and localization: A scalable
  real-time causal approach}.
\newblock International Journal of Robotics Research (2011)

\bibitem{Kelly:2011bw}
Kelly, J., Sukhatme, G.S.:
\newblock {Visual-Inertial Sensor Fusion: Localization, Mapping and
  Sensor-to-Sensor Self-calibration}.
\newblock International Journal of Robotics Research (2011)

\bibitem{Mourikis:2007dm}
Mourikis, A.I., Roumeliotis, S.I.:
\newblock {A Multi-State Constraint Kalman Filter for Vision-aided Inertial
  Navigation}.
\newblock In: International Conference on Robotics and Automation, IEEE (2007)

\bibitem{Li:2013co}
Li, M., Mourikis, A.I.:
\newblock {High-precision, consistent EKF-based visual{\textendash}inertial
  odometry}.
\newblock International Journal of Robotics Research (2013)

\bibitem{Hesch:2013jf}
Hesch, J.A., Kottas, D.G., Bowman, S.L., Roumeliotis, S.I.:
\newblock {Towards Consistent Vision-Aided Inertial Navigation}.
\newblock Algorithmic Foundations of Robotics (2013)

\bibitem{Civera:2009kc}
Civera, J., Bueno, D.R., Davison, A.J., Montiel, J.M.M.:
\newblock {Camera self-calibration for sequential Bayesian structure from
  motion}.
\newblock In: International Conference on Robotics and Automation, IEEE (2009)

\bibitem{Li:2014jx}
Li, M., Yu, H., Zheng, X., Mourikis, A.I.:
\newblock {High-fidelity sensor modeling and self-calibration in vision-aided
  inertial navigation.}
\newblock In: International Conference on Robotics and Automation, IEEE (2014)

\bibitem{sturm1999plane}
Sturm, P.F., Maybank, S.J.:
\newblock On plane-based camera calibration: A general algorithm,
  singularities, applications.
\newblock In: Computer Vision and Pattern Recognition, 1999. IEEE Computer
  Society Conference on., IEEE (1999)

\bibitem{zhang2004extrinsic}
Zhang, Q., Pless, R.:
\newblock Extrinsic calibration of a camera and laser range finder (improves
  camera calibration).
\newblock In: Intelligent Robots and Systems, 2004.(IROS 2004). Proceedings.
  2004 IEEE/RSJ International Conference on, IEEE (2004)

\bibitem{martinelli2006automatic}
Martinelli, A., Scaramuzza, D., Siegwart, R.:
\newblock Automatic self-calibration of a vision system during robot motion.
\newblock In: International Conference on Robotics and Automation (ICRA), IEEE
  (2006)

\bibitem{kummerle2011simultaneous}
K{\"u}mmerle, R., Grisetti, G., Burgard, W.:
\newblock Simultaneous calibration, localization, and mapping.
\newblock In: Intelligent Robots and Systems (IROS), 2011 IEEE/RSJ
  International Conference on, IEEE (2011)

\bibitem{strasdat2010real}
Strasdat, H., Montiel, J., Davison, A.J.:
\newblock Real-time monocular slam: Why filter?
\newblock In: Robotics and Automation (ICRA), 2010 IEEE International
  Conference on, IEEE (2010)

\bibitem{brookshire2013extrinsic}
Brookshire, J., Teller, S.:
\newblock Extrinsic calibration from per-sensor egomotion.
\newblock Robotics: Science and Systems VIII, Jul (2013)

\bibitem{levinson2014unsupervised}
Levinson, J., Thrun, S.:
\newblock Unsupervised calibration for multi-beam lasers.
\newblock In: Experimental Robotics, Springer (2014)

\bibitem{sheehan2012self}
Sheehan, M., Harrison, A., Newman, P.:
\newblock Self-calibration for a 3d laser.
\newblock The International Journal of Robotics Research (2012)

\bibitem{Keivan:2014bl}
Keivan, N., Sibley, G.:
\newblock {Constant-time monocular self-calibration}.
\newblock Robotics and Biomimetics (ROBIO) (2014)  1590--1595

\bibitem{jauffret2007observability}
Jauffret, C.:
\newblock Observability and fisher information matrix in nonlinear regression.
\newblock IEEE Transactions on Aerospace and Electronic Systems (2007)

\bibitem{hermann1977nonlinear}
Hermann, R., Krener, A.:
\newblock Nonlinear controllability and observability.
\newblock IEEE Transactions on automatic control (1977)

\bibitem{hansen1998rank}
Hansen, P.C.:
\newblock Rank-deficient and discrete ill-posed problems: numerical aspects of
  linear inversion.
\newblock SIAM (1998)

\end{thebibliography}
\end{document}